\documentclass{article}

\PassOptionsToPackage{sort&compress, numbers}{natbib}
     \usepackage[final]{neurips_2019}
    
\usepackage[utf8]{inputenc} 
\usepackage[T1]{fontenc}    
\usepackage{hyperref}       
\usepackage{url}            
\usepackage{booktabs}       
\usepackage{amsfonts}       
\usepackage{nicefrac}       
\usepackage{microtype}      
\usepackage{amsmath}

\title{Missingness as Stability: Understanding the Structure of Missingness in Longitudinal EHR data and its Impact on Reinforcement Learning in Healthcare}

\author{%
  Scott L. Fleming$^{1}$, Kuhan Jeyapragasan$^2$, Tony Duan$^2$, Daisy Ding$^1$, \\ \textbf{Saurabh Gombar$^1$, Nigam Shah$^1$, Emma Brunskill$^2$} \\
  $^1$Stanford School of Medicine\\
  $^2$Department of Computer Science, Stanford University\\
  \texttt{scottyf@stanford.edu} \\
}

\begin{document}

\maketitle

\begin{abstract}
There is an emerging trend in the reinforcement learning for healthcare literature. In order to prepare longitudinal, irregularly sampled, clinical datasets for reinforcement learning algorithms, many researchers will resample the time series data to short, regular intervals and use last-observation-carried-forward (LOCF) imputation to fill in these gaps. Typically, they will not maintain any explicit information about which values were imputed. In this work, we (1) call attention to this practice and discuss its potential implications; (2) propose an alternative representation of the patient state that addresses some of these issues; 
and (3) demonstrate in a novel but representative clinical dataset that our alternative representation yields consistently better results for achieving optimal control, as measured by off-policy policy evaluation, compared to representations that do not incorporate missingness information.
\end{abstract}

\section{Introduction}

Recent work in reinforcement learning (RL) for health-related problems has highlighted some of the promises and pitfalls of the RL approach for clinical data \cite{komorowski2018artificial, jeter2019does, saria2018individualized, peng2018improving, gottesman2018evaluating, yu2019reinforcement}. An important but perhaps understated fact throughout this literature is that how one models the patient health state matters: policies learned under one patient state representation may differ dramatically from another, even though the two policies are ultimately learned from the same underlying raw data \cite{gottesman2019guidelines}. An added difficulty is that traditional methods in reinforcement learning are not well-adapted to missingness and the irregular sampling frequencies so commonly observed in clinical data \cite{pivovarov2014identifying, sutton2018reinforcement}. 
RL methods offer a promising framework for optimizing sequences of treatment decisions in the clinic, but these methods will only succeed if they have access to adequate representations of the patient state.

One emerging trend in the recent offline reinforcement learning for healthcare literature is to preprocess the clinical data by discretizing and resampling observations into regular time intervals using ``Last-Observation-Carried-Forward'' (LOCF) imputation \cite{data2016secondary}. In this regime, one substitutes a missing value (e.g. heart rate) with the last observed value for that variable \cite{gelman2006data}. However, few of the reinforcement learning for healthcare papers which utilize LOCF imputation 
mention any use of 
methods 
to leverage missingness information \cite{Jagannatha2018, lopez2019deep, nemati2016optimal, prasad2017reinforcement}. Here, we explore the impact of this choice and potential alternatives. Our results suggest that (1) dynamics of the patient health state differ significantly between periods of more frequent missingness and less frequent missingness, and (2) incorporating missingness information into the patient state representation can improve policy performance, as measured by off-policy policy evaluation (OPPE). 

\section{Background and Related Work}

While researchers have rigorously analyzed the theory of missing data and its implications \cite{rubin1976inference, roderick2002statistical}, missingness in electronic health record (EHR) data specifically is less well studied. 
Early indications suggest, however, that missingness information is a critical component of any adequate patient time series representation. 
This may be because missingness can provide insight into the true (but unobserved) value of the missing entity itself as well as more general information about the patient's health state. Supporting this idea, \cite{beaulieu2018characterizing} explored patterns of missingness in an EHR dataset and found that a substantial portion of common laboratory values appeared \emph{not} to be Missing Completely at Random (where 
observed/missing data would have the same distribution), but rather Missing at Random (where systematic differences between the missing and observed values can occur when considered in isolation, but these differences are fully explained by other observed variables). 
This has important implications. 
\cite{che2018recurrent} found that rates of missingness for many variables in the MIMIC-III dataset were highly correlated with ICD-9 and diagnosis categories. 
\cite{afessa2005influence} found that patients with higher rates of missingness in variables related to an Acute Physiology Score (APS) experienced lower mortality rates compared to those with lower rates of missingness, even after controlling for patients' severity of illness. 

Why might missing values appear in a patient time series extracted from the EHR, and how might they give rise to the above phenomena? We suggest at least two potential mechanisms for missingness, each informative in its own way. The first essentially arises as an artifact of binning a multidimensional time series into discrete windows: if one laboratory test is administered every 24 hours, another every 6 hours, and we bin the patient time series into 4-hour windows, then inevitably some windows will have missing entries for the two tests. Note, however, that if a laboratory test is administered more frequently in sick individuals than in healthy individuals, then rates of missingness for the test will be higher in healthy (less frequently tested) individuals under a fixed binning scheme. Missingness in this context may therefore be a proxy for testing frequency, which in turn can be informative of a patient's overall health. An illustrative example where more frequent sampling (and, consequently, lower rates of missingness in the medical record) is associated with poorer health is given by \cite{duff2018frequency}. 

A second mechanism of missingness arises when a particular laboratory test is only administered (or not administered) if the ordering physician has a suspicion about some aspect of the patient's health. For example, ICU doctors will often order an erythrocyte sedimentation rate (ESR) test on a patient for whom they suspect inflammation or infection \cite{fincher1986clinical, brigden1999clinical}. The \emph{absence} of an ESR value in the EHR, then, could indicate that the clinician did \emph{not} suspect infection or inflammation. This gives additional information about the patient's health state. Illustrating how broad and systematic this mechanism of missignness can be in the EHR, \cite{agniel2018biases} found that just the presence of a laboratory test in a patient's EHR - completely ignoring its actual value - was significantly associated with odds of survival for 86\% of the 272 common clinical laboratory tests they analyzed.

In aggregate, 
current literature supports the notion of "informative missingness" \cite{che2018recurrent, rubin1976inference}, i.e. that missingness itself carries important information about both the value of the missing entity and the patient state. Given the recent attention on reinforcement learning methods as an approach for analyzing longitudinal EHR data, an important lingering question concerns how exactly one should handle missing values in these types of studies. Several approaches from the non-RL literature deserve consideration. \cite{che2018recurrent} developed a unique LSTM structure that incorporates information about a variable's missingness and time since last observation and demonstrated that their model improves predictive performance. \cite{agor2019value},  \cite{sharafoddini2019new}, and \cite{lipton2016directly} all demonstrate that even just the simple practice of including in the feature set an indicator to identify when imputed variables were originally missing can significantly improve performance of predictive models for critical patient outcomes. In this paper, for the sake of illustration and to encourage adoption in practice, we lean toward simple methods over relatively complex ones for incorporating missingness information into the patient state representation and therefore take the latter approach. To the best of our knowledge, ours is the first work to evaluate the impact of including a post-imputation indicator for missingness in the representation of the patient state for applications of reinforcement learning 
to healthcare data.

\section{Dataset}

We extracted EHR data from the STRIDE database \cite{lowe2009stride} for 8,983 ICU patients who underwent unfractionated heparin (UFH) anticoagulation therapy between 2012 and 2018. UFH is an important anticoagulant delivered to ICU patients who are at risk of developing venous thromboembolism (VTE) -- blood clots that can form in a patient’s veins and block the flow of blood to critical organs. In consultation with a clinical care team, all laboratory values deemed relevant to optimal UFH dosing decisions were included in our analysis. Observations in the dataset were defined based on the times at which a physician simultaneously requested an anti-Xa chromogenic assay (anti-Xa) and activated partial thromboplastin time (aPTT) laboratory tests, as these two tests are the standard protocol for UFH monitoring at Stanford Hospital. These orders are frequently made concurrently with a suite of other laboratory tests relevant to the task of UFH dosing, including those listed in Table \ref{mann_whitney}.

The example of UFH dosing is representative of a broader class of optimal control problems often encountered in clinical settings. If a doctor gives a patient too much UFH, they can put the patient at risk of excessive and dangerous bleeding. If the doctor does not give the patient enough UFH, they may increase the patient’s risk of developing a VTE \cite{hylek2003challenges}. For this reason, doctors typically try to titrate a patient’s UFH dosage so as to 
maintain the values of laboratory tests which measure clotting tendency, namely aPTT and anti-Xa, within target therapeutic ranges \cite{baglin2006guidelines}. Abnormalities in liver function, however, as measured by tests like ALT, AST, and TBILI (see Table \ref{mann_whitney}) may alter metabolism of UFH. This can lead to deceptive or unstable aPTT values \cite{vandiver2012antifactor} as well as other more serious complications \cite{price2013discordant}. For this reason, the presence of a lab value like TBILI in a patient's EHR may suggest that the patient is unstable or that the doctor is concerned about more serious underlying physiological issues. Vice versa, the absence of such a lab value may indicate relative stability.

\begin{table}
  \caption{Mann-Whitney U Test evaluating whether aPTT dynamics are significantly different under concurrent laboratory test missingness vs. non-missingness.  (\textbf{bold}: significant after family-wise error rate control 
  ). Only significant results are shown here; see Table \ref{mann_whitney_appendix} in Appendix for full results.}
  \label{mann_whitney}
  \centering
  \begin{tabular}{llll}
    \toprule
    \cmidrule(r){2-3}
    Concurrent Lab, $L$ & Median $X_{L, missing}$ & Median $X_{L, not \ missing}$ & $p$-value \\
    \midrule
    Prothrombin Time (PT)      & 27.10 & 35.44 & \textbf{1.61e-10} \\
    International Normalized Ratio (INR)     & 27.10 & 35.44 & \textbf{1.61e-10} \\
    Total Bilirubin (TBILI)   & 32.71 & 36.21 & \textbf{2.01e-4} \\
    Aspartate Aminotransferase (AST)     & 32.67 & 36.24 & \textbf{2.12e-4} \\
    Alanine Aminotransferase (ALT)     & 32.71 & 36.23 & \textbf{2.57e-4} \\
    Erythrocyte Sedimentation Rate (ESR)     & 34.99 & 39.13 & \textbf{2.20e-5} \\
    \bottomrule
  \end{tabular}
\end{table}

\section{Models and Methods}

\paragraph{Statistical testing.} Given that aPTT values in particular are known to be susceptible to aberrations in liver function, we first tested the hypothesis of whether the dynamics of aPTT, one of the lab values that clinicians would like to keep in control, are significantly different when one of the concurrently-measured laboratory tests is missing vs. not missing. In order to do this, we computed:
\begin{align}
    X_{L, missing} = \{| \textnormal{aPTT}_{t + 1} - \textnormal{aPTT}_{t - 1}| : L_{t}\ \mathrm{missing}\} \\
    X_{L, not\ missing} = \{| \textnormal{aPTT}_{t + 1} - \textnormal{aPTT}_{t - 1}| : \ L_{t}\ \mathrm{not}\ \mathrm{missing}\}
\end{align}

where $L_t$ represents the value of laboratory test $L$ (e.g. total bilirubin) corresponding to an observation taken at time $t$, $\textnormal{aPTT}_{t-1}$ represents the observed aPTT value one observation before this time point, and $\textnormal{aPTT}_{t + 1}$ represents the observed aPTT value one observation after. We then performed a one-sided Mann-Whitney U test \cite{mann1947test}, testing against the null hypothesis that a randomly drawn observation from $X_{L, missing}$ is equally likely to be less than or greater than a randomly drawn observation from $X_{L, not\ missing}$. 
Significant findings from this analysis (after controlling for multiple hypothesis tests with Bonferroni correction 
) can be found in Table \ref{mann_whitney}, with full results in the Appendix (Table \ref{mann_whitney_appendix}).


\paragraph{Reinforcement learning.}

We assessed the impact of including missingness information in the patient state representation on the performance of both Off-Policy Policy Optimization (OPPO) with Fitted Q-Iteration and Off-Policy Policy Evaluation (OPPE) with several Inverse Propensity Scoring (IPS) methods. We defined a reward at each step of the trajectory based on whether or not a patient's aPTT and anti-Xa values fell within what is considered to be a ``therapeutic range''. Both aPTT and anti-Xa measure the patient's blood's propensity to coagulate inappropriately (e.g. too slowly or too quickly). Clinical guidelines suggest that therapeutic ranges are 40 to 80 seconds for aPTT and 0.3 to 0.7 IU/mL for anti-Xa. In each step of a trajectory we define the reward as a sum of -1 for every observation irrespective of the control region (to penalize staying in the ICU), -1 if the patient was not in the therapeutic range for anti-Xa, and -1 if the patient was not in the therapeutic range for aPTT. 

We split our dataset into 5 folds (randomized by patient ID so that all of a patient's visits in the clinic were in the same fold) and performed cross-validation on our entire imputation, OPPO, and OPPE pipeline for each fold \cite{fushiki2011estimation, hastie2005elements}. Thus, for each of the 5 splits of the dataset into training and validation folds, we performed the following steps: (1) for every variable that exhibited missingness in the raw dataset, we added a binary indicator to the patient state representation whose value took on 1 if the corresponding variable was originally missing and 0 otherwise; (2) we performed LOCF imputation on all missing values in the dataset for which a prior value in the patient's trajectory was available; (3) for missing values with no prior value available, we used iterative multivariate imputation, learning the imputation model from the training folds and applying that model to the heldout validation fold \cite{buuren2010mice, buck1960method, scikit-learn}; (4) we learned an optimal policy from the training folds using Fitted Q-Iteration with an Extra-Trees regressor \cite{ernst2005tree}; (5) we evaluated that policy on the held-out validation fold using standard importance sampling (IS), standard weighted importance sampling (WIS), step-wise IS, and step-wise WIS  methods \cite{jiang2015doubly}. 
We used these Inverse Propensity Scoring (IPS) methods for OPPE because they have theoretical guarantees of consistency under assumptions of coverage and nonconfounding and are popular in OPPE literature \cite{Voloshin2019}. 
For each split of the dataset, we performed OPPO with and without the missingness information indicator followed by OPPE with and without the missingness indicator. We do not, however, report evaluation metrics for OPPO with a missingness indicator followed by OPPE without a missingness indicator because of potential confounding \cite{miettinen1974confounding}.

\begin{table}
  \caption{Impact of Including Missingness Information in Patient State Representation on Off-Policy Policy Optimization (OPPO) and Off-Policy Policy Evaluation (OPPE). Numbers reported are the average $\pm$ 1 standard deviation of OPPE estimates over the validation folds from 5-fold CV.}
  \label{ope_heparin}
  \centering
  \begin{tabular}{cccccc}
    \toprule
    \multicolumn{2}{c}{Missingness Information Used?} & \multicolumn{4}{c}{Expected Return (Off-Policy Evaluation)}                \\
    \cmidrule(r){1-2}\cmidrule(r){3-6} 
    OPPO & OPPE & Standard IS & Standard WIS & Stepwise IS & Stepwise WIS\\
    \cmidrule(r){1-2}\cmidrule(r){3-6} 
    No & No  & -0.547 $\pm$ 0.074 & -2.159 $\pm$ 0.236 & -1.137 $\pm$ 0.097 & -3.843 $\pm$ 0.156 \\
    
    No & Yes & -0.538 $\pm$ 0.076  & -2.155 $\pm$ 0.246 & -1.118 $\pm$ 0.100 & -3.838 $\pm$ 0.178\\
    
    
    Yes & Yes & \textbf{-0.504 $\pm$ 0.086}  & \textbf{-2.049 $\pm$ 0.253} & \textbf{-1.046 $\pm$ 0.108} & \textbf{-3.710 $\pm$ 0.232} \\
    \bottomrule
  \end{tabular}
\end{table}

\section{Results and Discussion}

Table \ref{mann_whitney} shows the results of our test for whether the dynamics of aPTT differs under missingness of concurrent laboratory values. 
For six laboratory tests we found that the absolute difference between aPTT values measured just before and just after an aPTT/anti-Xa observation was significantly less, on average, when the the concurrent lab test was missing compared to when it was not missing. Notably, half of the concurrent labs found to have this significant effect (namely TBILI, AST, ALT) are typically used to asses liver damage \cite{newsome2018guidelines}, which is known to disrupt normal aPTT dynamics \cite{ha2016anticoagulation}. These findings suggest that missingness may be a proxy for aPTT test stability over time, and could thus be important when learning to make optimal treatment decisions based on patient state dynamics. 

Table \ref{ope_heparin} shows the effect of including missingness information into the patient state space for OPPO and OPPE. As a comparison, the average discounted rewards under the clinician's data-generating policy was -7.656 $\pm$ 0.268. 
On average, incorporating missingness into the patient state space for both OPPO and OPPE improved the estimated reward under every IPS method analyzed.
The effects appear to be additive in the sense that incorporating missingness into the state representation for OPPE improved estimated performance, and additionally incorporating missingness into 
OPPO increased estimated performance even further. Although these findings were provocatively consistent across IPS methods, they were not significant, likely due to known issues of high variance in OPPE estimators \cite{thomas2015high}. In accordance with our discussion in Section 2 about "informative missingness", our results suggest that missingness in our dataset may carry important information about the patient's health state, and that leveraging this missingness information can potentially improve estimated reward under OPPO and OPPE. Researchers applying reinforcement learning to healthcare datasets may benefit from this simple augmentation to the patient state representation. The code used to run our analyses can be found at \url{github.com/scottfleming/rl-missingness}. 

\medskip

\bibliography{references}
\pagebreak
\section{Appendix}
\begin{table}[h]
  \caption{Mann-Whitney U Test evaluating whether aPTT dynamics are significantly different under concurrent laboratory test missingness vs. non-missingness.  (\textbf{bold}: significant after Bonferroni correction for controlling the family-wise error rate 
  )}
  \label{mann_whitney_appendix}
  \centering
  \begin{tabular}{llll}
    \toprule
    \cmidrule(r){2-3}
    Concurrent Lab, $L$ & Median $X_{L, missing}$ & Median $X_{L, not \ missing}$ & $p$-value \\
    \midrule
    Prothrombin Time (PT)      & 27.10 & 35.44 & \textbf{1.61e-10} \\
    International Normalized Ratio (INR)     & 27.10 & 35.44 & \textbf{1.61e-10} \\
    Total Bilirubin (TBILI)   & 32.71 & 36.21 & \textbf{2.01e-4} \\
    Creatinine (CR)      & 27.49 & 35.27 & 2.22e-2 \\
    Aspartate Aminotransferase (AST)     & 32.67 & 36.24 & \textbf{2.12e-4} \\
    Alanine Aminotransferase (ALT)     & 32.71 & 36.23 & \textbf{2.57e-4} \\
    Platelet Count (PLT)     & 42.48 & 35.18 & 0.988 \\
    C-reactive Protein (CRP)     & 35.15 & 35.53 & 0.965 \\
    Red Cell Distribution Width (RDW)     & 33.67 & 35.19 & 0.561 \\
    Hemoglobin (HGB)     & 40.14 & 35.18 & 0.372 \\
    White Blood Cell Count (WBC)     & 33.67 & 35.19 & 0.561 \\
    Fibrinogen (FBN)     & 34.27 & 39.84 & 0.324 \\
    Erythrocyte Sedimentation Rate (ESR)     & 34.99 & 39.13 & \textbf{2.20e-5} \\
    \bottomrule
  \end{tabular}
\end{table}

\end{document}